\documentclass[runningheads]{llncs}

\usepackage{multirow}
\usepackage{enumitem}
\newtheorem{ex}{Example}
\usepackage{thmtools}
\usepackage{nameref}
\usepackage[hidelinks]{hyperref}
\usepackage{graphicx}
\usepackage[nameinlink,noabbrev,capitalise]{cleveref}
\usepackage{ifthen}
\newboolean{review}
\setboolean{review}{false}

\begin{document}

\title{CO-Fun: A German Dataset on Company Outsourcing in Fund Prospectuses for Named Entity Recognition and Relation Extraction}
\titlerunning{CO-Fun: A German Dataset on Company Outsourcing in Fund Prospectuses}

\author{
	Neda Foroutan \and 
	Markus Schröder \and 
	Andreas Dengel
}
\institute{
	Smart Data \& Knowledge Services Dept., DFKI GmbH,\\Trippstadter Str. 122, 67663 Kaiserslautern, Germany\\
	\email{\{neda.foroutan,markus.schroeder,andreas.dengel\}@dfki.de}
}

\maketitle

\begin{abstract}
The process of cyber mapping gives insights in relationships among financial entities and service providers.
Centered around the outsourcing practices of companies within fund prospectuses in Germany, we introduce a dataset specifically designed for named entity recognition and relation extraction tasks.
The labeling process on $948$ sentences was carried out by three experts which yields to $5,969$ annotations for four entity types (Outsourcing, Company, Location and Software) and $4,102$ relation annotations (Outsourcing--Company, Company--Location).
State-of-the-art deep learning models were trained to recognize entities and extract relations showing first promising results. %
An anonymized version of the dataset, along with guidelines and the code used for model training, are publicly available at \url{https://www.dfki.uni-kl.de/cybermapping/data/CO-Fun-1.0-anonymized.zip}.

\end{abstract}

\section{Introduction}

Cyber incidents, such as data breaches and ransomware attacks, pose potential risks to financial stability since banks and other institutes increasingly outsource processes and services to information and communication technology providers \cite{Adelmann2020CyberRA}.
To discover cyber risks, a conceptual method is ``cyber mapping'' -- a process which links the financial network (e.g. banks, funds, insurance companies) with the cyber network (e.g. cloud services, datacenters, software providers) \cite{BrauchleGoebelSeiler}.
Evidences for constructing a mapping could be descriptions of outsourced services and companies in the financial domain.
A promising source to collect such hints can be found in publicly available fund prospectuses.
In these documents, German Capital Management Companies (CMCs) have to state outsourcing companies and their provided services for a particular fund. 
To give an example, consider the following simplified sentence.
\begin{ex}
\label{ex1}
Die Gesellschaft hat Rechenzentrumsleistungen auf die Mercurtainment \& CO KGaA ausgelagert.\\
`The company has outsourced data center services to Mercurtainment \& CO KGaA.'
\end{ex}
To extract structured information, a usual step in Natural Language Processing (NLP) is the application of Named Entity Recognition (NER) to discover entities in texts. 
In our scenario, there are outsourced services (e.g. ``data center services'') and companies (e.g. ``Mercurtainment \& CO KGaA'').
After that, Relation Extraction (RE) is commonly used to predict relationships between entities, in our case, services and companies.
In order to train such NLP models, a dataset with ground truth labels is necessary.

In this paper, we present a novel dataset to support the process of cyber mapping using NLP models.
Our annotated corpus consists of $948$ sentences extracted from $1,054$ German fund prospectuses.
In total, $5,969$ named entity annotations and $4,102$ relation annotations were added by experts to acquire ground truth data.
We conducted experiments with our dataset to evaluate the performance of trained models.

\section{Related Work}

NER and RE tasks are fundamental building blocks for extracting information within unstructured texts (for a recent survey see \cite{ReviewNerRE}). %
For training models, several corpora have been built to cover specific domains, for example, the biomedical area \cite{REcorpus} or for clinical purpose \cite{chilean}. %
Some of them targeting specific languages, like Kazakh \cite{kaznerd} and Italian \cite{kind}.
Regarding German language, \cite{GeoNerRe} collected data from tweets, news documents and RSS feeds to create a corpus with named entities such as Disasters, Triggers, Location, Organizations, Persons as well as 15 relations of the mobility and industry domain.

More related to our scenario is the business domain since the discovery of relationships between company entities is of interest.
Here, \cite{OrgProduct2020} provided an English dataset for recognizing companies, products and their relations to each other.
The data was gathered from company homepages, business news portals, forums and social media channels. %
Instead of considering the product in the business relation, extracting the relation between two companies within unstructured texts -- called Business Relation Extraction -- has attracted attention in research and industry.
\cite{multilevelREEng} presented a web-based English dataset for the business relation extraction between organizations. %
They also recommended a relation classifier using multilevel knowledge of entities to predict five types of relations between companies, i.e. Investment, Cooperation, Sale-purchase, Competition and Legal proceedings.
In subsequent work, the authors provided the BIZREL dataset \cite{BIZREL2022s}, a multilingual corpus in French, Spanish and Chinese in addition to their introduced English dataset.
Similarly, they collected data via keyword queries using well-known search engines and the same five types of relations.
\cite{AsyRelDir} proposed a method of iteratively extracting asymmetric business relations like ``owner-of'' between two companies and indicating the relation direction between them.
They evaluated their suggested method on two datasets based on \textit{New York Times} News articles.
In the financial domain, \cite{FinFrench} created a French corpus including 26 entity types and 12 relation types gathered from French financial newspapers.
They trained a BERT-based \cite{devlin2019bert} NER model on five types of entities (Person, Location, Organization, Role and Currency) and investigated a rule-based RE method for the relationship around the Role entity (i.e. ``hasRole''). 
Moreover, \cite{kpi-bert} recommended a BERT-based architecture that employs a Gated Recurrent Units tagger coupled with conditional label masking to jointly predict entities tags sequentially and links the predicted entities. 
Additionally, they built a dataset from real-world German financial documents.
The main entity type is Key Performance Indicators (KPI), such as revenue or interest expenses.
Generally, entity classes include KPI, change of it, its monetary value and their sub-types.
Linked relations are considered between KPI and sub-types or their values.

Still, there seems to be no dataset which meets our requirements.
To train NLP models for performing cyber mapping, we need realistic sentences in German language explicitly mentioning outsourced services in the financial domain.
Therefore, we built our own dataset from fund prospectuses which is covered in detail in the next section.

\section{Corpus Creation}

The corpus was created in a collaborative research lab of Deutsche Bundesbank\footnote{\url{https://www.bundesbank.de/en}} (the central bank of the Federal Republic of Germany) and
\ifthenelse{\boolean{review}}{the paper's authors\footnote{Anonymized for double-blind review.}}{the German Research Center for Artificial Intelligence\footnote{\url{https://www.dfki.de/}} (DFKI)}.
In this project a set of $1,054$ publicly available fund prospectuses (PDFs) were collected from websites of $37$ well-known Capital Management Companies (CMCs) in Germany.
Our corpus is built upon these documents by first converting the PDFs into plain texts using Apache's PDFBox\footnote{\url{https://pdfbox.apache.org/}} text stripper routine.
The fund prospectuses consist of $92$ pages on average, however, only a certain section in the document, usually no longer than a full or half page, mentions outsourced services.
Conveniently, independent of the CMC, such a section is commonly named `Auslagerung [Outsourcing]' followed by a section labeled `Interessenkonflikte [Conflicts of Interest]' with some minor variations.
Therefore, with a proper regular expression we were able to identify the beginning and end of these sections in our plain texts.
For sentence splitting, Apache's OpenNLP\footnote{\url{https://opennlp.apache.org/}} sentence detector loaded with a German pre-trained model\footnote{\href{https://opennlp.apache.org/models.html}{\texttt{opennlp-de-ud-gsd-sentence-1.0-1.9.3.bin}}} was applied.
To turn words with hyphens such as `Dienst--leistung [ser–vice]' in their hyphenless form, string matching and string manipulation was performed with regular expressions.
Finally, $1,267$ unique sentences could be collected of comparable shape as Example~\ref{ex1}.
However, roughly half of them assemble bullet point lists.

\subsection{Annotation Process}

\begin{figure}[!ht]
    \begin{center}
        \includegraphics[width=\linewidth]{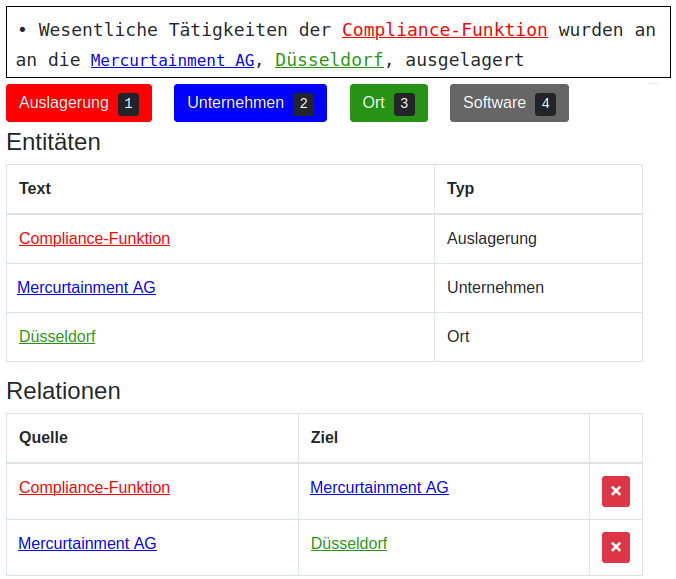}
        \caption{A graphical user interface in German to annotate a sentence (top) with named entities (center) and relations (bottom). Entity types are `Auslagerung [Outsourcing]', `Unternehmen [Company]', `Ort [Location]' and Software.}
        \label{fig:annotation-gui}
    \end{center}
\end{figure}

Three subject-matter experts of the Deutsche Bundesbank [German Federal Bank] annotated the corpus with named entities and relations.
For this, a Graphical User Interface (GUI) was provided which is depicted in  \cref{fig:annotation-gui}.
Sentences are randomly distributed to the annotators who independently worked on them.
Because of limited time available, same texts were not sent to multiple annotators, therefore, expert agreement is not considered.
The following named entity types could be annotated: `Auslagerung' (Engl.: Outsourcing), `Unternehmen' (Engl.: Company), `Ort' (Engl.: Location) and Software.
Additionally, the GUI allows users via drag and drop operations to declare the following two relationships: Outsourcing--Company and Company--Location.
Annotators could mark sentences as ignorable if they recognize that no entities are present.
To reduce annotation efforts, our system pre-annotates sentences with already collected named entities once they exactly match in the text.
A three-page annotation guideline was provided to give a brief tutorial and to cover special cases during the annotation process.
The experts sporadically annotated sentences during their working hours and were done after around one month.

\subsection{Resulting Dataset}

Our Company Outsourcing in Fund Prospectuses (CO-Fun) dataset consists of $948$ sentences with $5,969$ named entity annotations, including $2,340$ Outsourced Services, $2,024$ Companies, $1,594$ Locations and $11$ Software annotations.
While the sentences have an average length of $314.8 \pm 393.7$\footnote{using $\pm$ notation for standard deviation} characters, they contain $6.3 \pm 9$ annotations on average.
Without considering duplicates, our corpus mentions $270$ outsourced services, $323$ companies, $84$ locations and one software.
Regarding relations, we have found $2,573$ Outsourcing-Company relationships and $1,529$ links between Companies and Locations (in total $4,102$).
On average, $4.3 \pm 8.6$ relations can be found in the sentences.

The raw data of CO-Fun consists of records formatted in JavaScript Object Notation\footnote{\url{https://www.json.org/}} (JSON) which are sequenced in a JSON-line file (UTF-8 encoding).
Each entry has the following properties:
In the \texttt{text} property, the annotated text is present in form of HyperText Markup Language (HTML).
We use \texttt{span} tags\footnote{\url{https://html.spec.whatwg.org/\#the-span-element}} to annotate named entities in text.
Annotations are uniquely identified with a Universally Unique Identifier (UUID) (\texttt{id}).
The entity's type is given in the \texttt{type} attribute which can be one of the following options: 
`Auslagerung' (Engl.: Outsourcing), `Unternehmen' (Engl.: Company), `Ort' (Engl.: Location) or Software.
Additionally, named entities are listed in a JSON array called \texttt{entities}, again with their ID, type and covered text.
Another JSON array (\texttt{relations}) defines the relationships between a source entity (\texttt{src}) and a target entity (\texttt{trg}).
Gathering a dataset about outsourced services to German companies naturally raises concerns of potential misuse.
We therefore performed an anonymization of all companies by randomly  swapping their names with other companies. %
For this, we make use of OffeneRegister\footnote{\url{https://offeneregister.de/}} -- a database dump of the German commercial register.

The anonymized CO-Fun dataset is publicly available\footnote{\url{https://www.dfki.uni-kl.de/cybermapping/data/CO-Fun-1.0-anonymized.zip}} under MIT license together with other related materials such as the annotation guideline, derived data and source code.
In the next section, initial experiments with our dataset are presented.

\section{Experiments}
In our study, we investigated extracting two types of structured information from our corpus. 
Firstly, we recognized entities within our sentences by applying Named Entity Recognition (NER) methods. 
Secondly, we detected relations between entities using a Relation Extraction (RE) model.

\subsection{NER Methods}
We employed two NER models: Conditional Random Fields (CRF) and BERT (Bidirectional Encoder Representations from Transformer) \cite{devlin2019bert}. 
For applying CRF, we utilized CRFsuit toolkit \cite{okazaki2007crfsuite} and derived the features related to the token itself and its neighborhood information.
The token features include the word itself, its part-of-speech tag, whether the word is capitalized, starts with a capital letter or is a digit.
Additionally, we considered the bigram and trigram characters the word ends with, and each token was assigned the same bias feature.
Furthermore, we captured neighborhood information from the two words to the left and right of the token, checking their part-of-speech tags and if they start with a capital letter or are entirely in uppercase.
If a token is at the beginning or end of the sentence, we provided BOS or EOS as the left or right neighbor to CRF, respectively.

In order to apply the pre-trained BERT model, we fine-tuned the German language version of it on our data using the SpaCy 3 library\footnote{\url{https://spacy.io/}}.
As a result, the model with about 110 million parameters is capable of predicting our four entity types. %

\subsection{NER Dataset}
Before applying the CRF model, each sentence was tokenized and an Inside-Outside-Beginning (IOB) format label was assigned to each token.
The IOB scheme gives each token one of the following labels: B-ent, I-ent or O.
If the token is the beginning of an entity, it is labeled as B-ent (begin-of-entity) but if the token is part of the entity but not its beginning, I-ent (inside-of-entity) is assigned to the token. 
If the token does not belong to any of the entity types, it is tagged as 'O'.
After tagging the tokens of each sentence, in order to create the training, development and test sets, we randomly split the data with the proportion of 80\%, 10\% and 10\%, respectively.

For the BERT model, the same sentences were considered in each of three sets.
Each set includes a list of sentences with the list of tuples containing their entities and labels specified with the location of the entity in the sentence (start and end character position as well as entity label).
For later reuse, the training, development and test sets were converted into SpaCy binary files.

\subsection{RE Method}
The RoBERTa model \cite{DBLP:journals/corr/abs-1907-11692} is an optimized BERT model with approximately 355 million parameters. 
As a basis, we used SpaCy's tutorial for a relation extraction component on GitHub\footnote{\url{https://github.com/explosion/projects/tree/v3/tutorials/rel\_component}}.
In this project, the pre-trained RoBERTa model with the configuration of the German language is fine-tuned to extract relations.
Moreover, there is a max-length parameter representing the furthest distance at which existing relation is sought between any two entities.
We discovered in tests that the model performed best with a max-length of 20.

\subsection{RE Dataset}
The dataset split from the NER case is the same for the RE datasets.
For each sentence, a list of entities and relations were prepared.
A structure is provided for each entity to record an entity's text and label as well as its character and token position in text.
Each relation entry has a label and refers to a child and head entity using their token positions.
Ultimately, dataset text files were converted to binary files in SpaCy format.

\subsection{NER and RE Results}
The CRF model was run for $100$ iterations using the L-BFGS training algorithm.
The L1 and L2 regularization terms tuned by using cross validation are $0.05$ and $0.01$, respectively.
Default values were used for the remaining hyperparameters provided by the CRFsuite toolkit.
The BERT model was fine-tuned on the German training set for $299$ epochs with the batch size of $128$.
The initial learning rate and warm-up step were set to $5*10^{-5}$ and $250$, respectively.
Also, the L2 weight decay rate with value of $0.01$ was applied.
Similarly, the RoBERTa model was fine-tuned for $52$ epochs with a $1,000$ batch size.
Remaining parameters were configured the same as in BERT's configuration.
Both models were trained on a NVIDIA RTX A6000 GPU which took 40 minutes (NER) and 9 minutes (RE). %

We evaluated the performance of our models in terms of exact match using precision, recall and F1-score \cite{metrics}.
\Cref{tab:NER} demonstrates the performance of the NER models on the training and test sets of CO-Fun, measured by micro-averaging. 
Both models of CRF and BERT face overfitting as test F1-scores show lower scores than their training values.

\begin{table}[!ht]
\begin{center}
\begin{tabular}{ | c | c c c | c c c | }
\hline
\multirow{2}{4em}{Models} & \multicolumn{3}{c|}{Train} & \multicolumn{3}{c|}{Test}\\
&P & R & F1 & P & R & F1\\
\hline
CRF & 96.7 & 95.1 & 95.9 & 95.7 & 93.0 & 94.3\\
BERT & 99.8 & 94.2 & 97.0 & 92.9 & 91.5 & 92.2\\
\hline
\end{tabular}
\end{center}
\caption{\label{tab:NER} Precision (P), Recall (R) and F1-score results of NER models on the training and test sets of CO-Fun.}
\end{table}
However, CRF performs better than BERT on the test set with F1-score of $94\%$.
Furthermore, RoBERTa could classify $86.35\%$ of the relations that exist between entities in the test set, as shown in \cref{tab:RE}.
\begin{table}[!ht]
\begin{center}
\begin{tabular}{ |c | c c c | c c c| }
\hline
\multirow{2}{4em}{Models} & \multicolumn{3}{c|}{Train} & \multicolumn{3}{c|}{Test}\\
&P & R & F1 & P & R & F1\\
\hline
RoBERTa & 89.4 & 81.7 & 85.3 & 86.5 & 86.1 & 86.3\\     
\hline
\end{tabular}
\end{center}
\caption{\label{tab:RE} Precision (P), Recall (R) and F1-score results of RE models on the training and test sets of CO-Fun.}
\end{table}

\section{Conclusion and Future Work}
In this paper, we introduced an annotated German dataset called CO-Fun which is a NER and RE dataset on company outsourcing in fund prospectuses.
Our dataset contains $948$ sentences with $5,969$ named entity annotations (including Outsourced Services, Companies, Location and Software) and $4,102$ annotated relations (Outsourcing--Company and Company--Location).
Applying state-of-the-art NER and RE models showed promising performances on CO-Fun.

In the future, we aim to extend this dataset with similar data and improve the performance of applied models by using additional knowledge, for example, by incorporating knowledge graphs in the training process. %

\bibliographystyle{llncs}
\bibliography{paper}

\end{document}